\documentclass[runningheads]{llncs}
\usepackage[T1]{fontenc}
\usepackage{enumitem}
\usepackage{graphicx}
\usepackage{amsmath}
\usepackage{amssymb}
\usepackage{booktabs} 
\usepackage{wrapfig}
\usepackage[misc]{ifsym}
\usepackage{hyperref}
	
\begin{document}
\title{SportsGPT: An LLM-driven Framework for Interpretable Sports Motion Assessment and Training Guidance}
\titlerunning{SportsGPT}
%
\author{Wenbo Tian\inst{1,2}\and
Ruting Lin\inst{3} \and
Hongxian Zheng\inst{3} \and
Yaodong Yang\inst{4} \and
Geng Wu\inst{2}\and
Zihao Zhang\inst{5} \and
Zhang Zhang\inst{1,2}\textsuperscript{(\Letter)}}
\authorrunning{W. Tian et al.}
%
\institute{School of Artificial Intelligence, UCAS, Beijing, China\and
MAIS, NLPR, Institute of Automation, CAS, Beijing, China\and
College of Education, Beijing Sport University, Beijing, China\and
KuTi Sports Technology, Shaanxi, China\and
Haoxiong Technology, Shanghai, China\\
\email{zzhang@nlpr.ia.ac.cn}}
\maketitle              
\begin{abstract}
Existing intelligent sports analysis systems mainly focus on "scoring and visualization," often lacking automatic performance diagnosis and interpretable training guidance. Recent advances in Large Language Models (LLMs) and motion analysis techniques provide new opportunities to address the above limitations. In this paper, we propose SportsGPT, an LLM-driven framework for interpretable sports motion assessment and training guidance, which establishes a closed loop from motion time-series input to professional training guidance. First, given a set of high-quality target models, we introduce MotionDTW, a two-stage time series alignment algorithm designed for accurate keyframe extraction from skeleton-based motion sequences. Subsequently, we design a Knowledge-based Interpretable Sports Motion Assessment Model (KISMAM) to obtain a set of interpretable assessment metrics (e.g., \textit{insufficient extension}) by contrasting the keyframes with the target models. Finally, we propose SportsRAG, a RAG-based training guidance model built upon Qwen3. Leveraging a 6B-token knowledge base, it prompts the LLM to generate professional training guidance by retrieving domain-specific QA pairs. Experimental results demonstrate that MotionDTW significantly outperforms traditional methods with lower temporal error and higher IoU scores. Furthermore, ablation studies validate the KISMAM and SportsRAG, confirming that SportsGPT surpasses general LLMs in diagnostic accuracy and professionalism.

\end{abstract}
\section{Introduction}
In recent years, artificial intelligence (AI) technologies, particularly computer vision and machine learning algorithms, are profoundly reshaping the fields of sports science. From fine-grained motion data acquisition through human pose estimation~\cite{1,2,3} , to intelligent analysis and visualization of team tactics~\cite{4,5,6,7} , and automated refereeing for scored events such as diving~\cite{8} and gymnastics~\cite{9}, AI techniques have brought unprecedented efficiency gains to competitive sports, significantly enhancing coaches' decision-making and athletes' training effectiveness.

Despite notable progress, current "intelligent" sports applications still face several critical bottlenecks. First, most systems remain limited to basic "scoring + visualization" functionalities and simple digitization pipelines~\cite{3,5,10,11,12,13,14}. Although these tools present kinematic and kinetic data efficiently, they lack automated diagnosis, reliable interpretation, and personalized training guidance. A simple scoring or scaling function is insufficient for athletes to identify the occurrence of incorrect motions or to understand how to rectify them. Moreover, many amateur athletes and coaches often lack the scientific expertise required to design systematic rehabilitation strategies. Traditional models and visualization tools fail to provide personalized and practical training guidance to improve the athlete's performance. Finally, low automation in critical stages-specifically keyframe extraction-severely bottlenecks system efficiency and precision. While keyframe extraction is essential for bridging raw video and biomechanical analysis~\cite{15,16,17,18}, existing works often rely on time-consuming manual annotation. Meanwhile, classic algorithms like Dynamic Time Warping (DTW)~\cite{19,20,21} struggle with inter-individual variability and speed disparities, resulting in poor robustness and a continued dependence on manual intervention.

To address the challenges, we propose SportsGPT, an LLM-driven framework for interpretable sports motion assessment and training guidance. For sports motion assessment, we develop a Knowledge-based Interpretable Sports Motion Assessment Model (KISMAM). Guided by professional domain knowledge, KISMAM compares user skeleton sequences with target motion models to generate a set of interpretable assessment metrics. To enable accurate motion assessment, we introduce MotionDTW, a two-stage time-series alignment algorithm that facilitates precise temporal alignment and computationally efficient keyframe localization, thereby allowing precise detection of erroneous movements. Finally, to facilitate effective intervention, we present SportsRAG, a RAG-based training guidance model that leverages a large-scale sports-specific knowledge base (including 200 authoritative textbooks, 50,000 expert QA pairs, and 1,000 user reports). SportsRAG uses the assessment vector as a query to retrieve domain-specific prompts, empowering general-purpose LLMs to generate scientific sports training and rehabilitation guidance.

Benchmark on Athletics Motion Analysis. The dataset comprises 390 video sequences covering five distinct motion categories: the standing long jump and four phases of the 100m sprint (Start, Acceleration, Mid-course, and Late). Each sequence is annotated with skeletal keypoints and temporal keyframes. Furthermore, we establish target motion models using 100 sequences from elite athletes to facilitate precise capability evaluation.

Extensive experiments validate the efficacy of our framework. The proposed MotionDTW achieves state-of-the-art alignment precision with average temporal errors of 0.91 and 1.59 frames on the 100m sprint (Acceleration phase) and standing long jump, respectively, while maintaining superior computational efficiency and stability to baselines. In downstream assessment tasks, our method yields diagnostic results with significantly higher Intersection over Union (IoU) with ground truth compare to standard DTW-based approaches. Furthermore, human evaluations demonstrate that SportsGPT surpasses general-purpose LLMs in accuracy, comprehensiveness, professionalism, and feasibility, highlighting its practical value for professional sports training.

The main contributions of this paper are summarized as follows:
\begin{itemize}[label=--, leftmargin=*, nosep]
	\item We propose \textbf{SportsGPT}, an LLM-driven framework for interpretable sports motion assessment and training guidance, realizing an intelligent closed loop from motion time-series input to professional training guidance.
    \item We introduce a unified solution comprising \textbf{MotionDTW}, a two-stage time series alignment algorithm, and the \textbf{KISMAM}, which achieves a probabilistic mapping from quantitative biomechanical features to interpretable motion assessment metrics.
    \item We develop \textbf{SportsRAG}, which integrates a massive sport-specific knowledge base (over 50k expert entries) with RAG techniques to transform quantitative biomechanical characteristics into expert-level actionable guidance.
\end{itemize}

\section{Related Work}
\subsubsection{Sports Visualization and Digital Analysis Systems.} 
Computer vision has established a robust foundation for sports motion acquisition. At the perception layer, object tracking captures real-time trajectories in team sports (e.g., soccer, basketball)~\cite{4,5,23}, while human pose estimation~\cite{24} and semantic segmentation~\cite{25} enable non-contact biomechanical extraction and technique observation. Building on these, sports visualization systems-widely adopted in diving, tennis, and figure skating~\cite{26}-digitize transient movements into analyzable visual data, such as postural discrepancies and trajectories.

However, a substantial interpretability gap remains. Current systems primarily present raw data (e.g., complex joint-angle curves, 3D plots) rather than actionable training plans, rendering the information unintelligible to general users without professional interpretation. Consequently, bridging the divide between basic digitization and interpretable semantic guidance-transforming "visualization" into "training plans"-remains a pivotal challenge for sports AI.
\subsubsection{Action Keyframe Extraction.}
In sports technical analysis and rehabilitation, keyframes bridge continuous motion data and quantitative evaluation. Rather than analyzing entire sequences, researchers focus on moments of abrupt mechanical change, extrema, or major energy transfer. DTW initially developed for speech recognition, is widely used for temporal alignment in ECG diagnosis~\cite{27}, gait recognition~\cite{27,28,29,30}, health monitoring~\cite{16,31,32}, and action recognition~\cite{23,33,34}, enabling precise sequence alignment and accurate keyframe extraction.

In rehabilitation, keyframes underpin real-time feedback and objective assessment.Automated Functional Movement Screening (FMS)~\cite{33} leverages manifold learning and DTW to identify critical positions like start, end, and lowest squat, supporting objective scoring.In competitive sports, keyframes facilitate technical evaluation. Golf studies~\cite{15,18} analyze phases such as address, top, and impact to compare trunk inclination and pelvic rotation, with Derivative DTW refining swing segmentation. In table tennis~\cite{35}, backswing completion and impact frames reveal joint-angle patterns linked to racket speed. 

Keyframes reduce the impact of inter-athlete variability in rhythm, speed, and duration, enabling precise phase-specific analysis. By discarding redundant frames while retaining essential biomechanical information information, keyframe extraction also improves computational efficiency and supports real-time feedback on resource-limited devices.
\subsubsection{Reference Models.}Reference models facilitate precise diagnosis by establishing kinematic baselines, thereby reducing individual variability. In competitive sports, elite athlete data define optimal solutions; approaches in sprinting~\cite{16} and golf~\cite{15} align learners against champion trajectories or professional templates (via DTW) to quantify technical deviations. Conversely, rehabilitation and skill acquisition rely on normative ranges and stratification. Studies compare subjects against age-matched controls or high-level groups~\cite{15,35} (e.g., in table tennis) to identify abnormalities and set progressive improvement goals, ensuring comparisons remain realistic and biologically valid.
\subsubsection{Domain-specific LLMs.}General LLMs, despite strong NLP performance, struggle in sports science and rehabilitation due to limited domain knowledge and tendency to "hallucinate" on biomechanics data or personalized plans~\cite{36}. This motivates vertical models pre-trained or fine-tuned on domain-specific data, often with multi-modal inputs.

In personal health, PH-LLM~\cite{31} processes time-series physiological data from wearables and, trained on 857 expert cases, delivers personalized sleep and fitness recommendations comparable to human experts. In competitive sports, multi-modal LLMs improve scoring and aesthetic analysis: Wang et al.~\cite{10} align audio with video in figure skating to segment key actions, generating technical scores and textual commentary like professional judges. In rehabilitation, foundation models enhance few-shot generalization; Mesquita et al.~\cite{32} show pre-trained time-series models outperform task-specific ones (AUC=73\%), enabling frame-level motion evaluation from limited video labels.

However, generative AI still falls short in exercise prescription. Puce et al.~\cite{37} note that models like ChatGPT-4 are adequate for basic guidance but lack load management, periodization, and real-time adaptability, producing conservative, formulaic recommendations. This highlights the need for domain-specific LLMs, which integrate biomechanics, provide interpretable diagnostics, and complement expert knowledge.

\section{Our Methods}
\begin{figure}[!ht]  
	\centering
	\includegraphics[width=\textwidth]{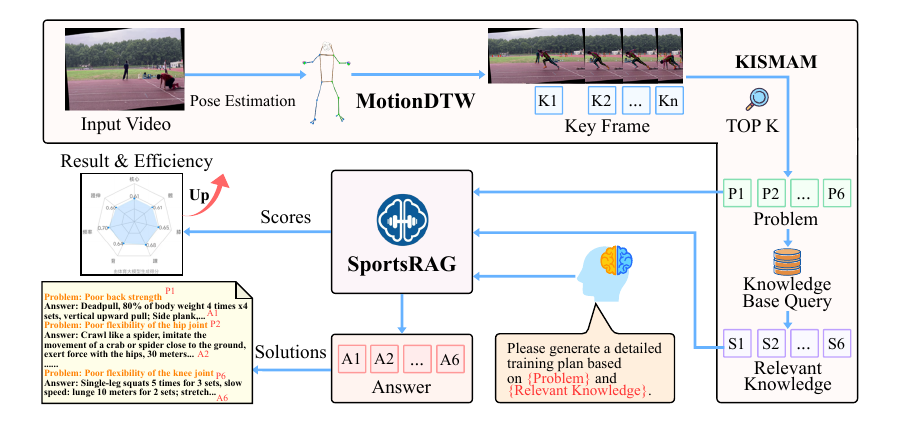}
	\caption{Architecture of the SportsGPT.} 
	\label{fig:framework}
\end{figure}

As illustrated in Figure~\ref{fig:framework}, the framework establishes a coherent closed-loop pipeline that transforms raw motion video into actionable expert guidance. The architecture proceeds through three stages. Initially, the Pose Estimation Module extracts skeletal coordinates from the input video. To mitigate temporal variations, MotionDTW employs a two-stage alignment algorithm. It aligns the user's sequence with a standard template to precisely extract critical keyframes ($K_1, \dots, K_n$), filtering out redundant transitional frames. Then, the KISMAM serves as the biomechanical reasoning engine. It contrasts the extracted user keyframes against professional target models. Through knowledge rule-based mapping, it converts kinematic deviations into a structured set of assessment metrics (visualized as a problem set $\{P_1, \dots, P_6\}$), semantically defining specific motion deficits. Finally, SportsRAG bridges the gap between diagnosis and prescription. Utilizing the assessment metrics as a semantic query, it retrieves verified intervention protocols ($\{S_1, \dots, S_6\}$) from a domain-specific Knowledge Base via RAG~\cite{22} techniques. These retrieved contexts guide the LLM to generate scored evaluations and precise, scientifically grounded training and rehabilitation guidance.

\subsection{Motion Evaluation module}
\subsubsection{MotionDTW~\cite{21}.}Traditional DTW applied to sports analysis often relies on raw coordinates or single-modal features, making it sensitive to athlete anthropometry, execution speed, and background noise, which limits keyframe extraction accuracy. To overcome this, we propose MotionDTW, which constructs a multi-modal feature space incorporating joint angles, angular velocities, and angular accelerations. Furthermore, it leverages temporal window contexts and joint weighting in a two-stage matching strategy for precise keyframe localization in long video sequences.

To reduce individual differences (e.g., height, limb length), MotionDTW uses biomechanical angular features instead of raw coordinates, ensuring invariance to in-plane rotation and translation. For each frame $t$, geometric angles of key joints (hip, knee, ankle, shoulder) are computed from three joint nodes $A, B, C$ (with $B$ as vertex). To capture dynamic characteristics, we include angular velocity and acceleration. For an angle sequence $\Theta = {\theta_1, \dots, \theta_T}$, derivatives are computed via central difference. Static and dynamic features are fused with weighting coefficients $\alpha$ and $\beta$ to form the multi-modal feature vector:

\begin{equation}
	F_{raw}^t = [\Theta_t, \alpha \cdot \dot{\Theta}_t, \beta \cdot \ddot{\Theta}_t]
\end{equation}where $\alpha=0.39$ and $\beta=0.25$ in our implementation, emphasizing velocity features for action phase representation.

To enhance robustness against local noise and capture short-term action trends, we apply a sliding window mechanism. For time $t$, features from the preceding and succeeding $w$ frames ($w=4$) are concatenated to form the temporal window feature. This enriches contextual information and reduces single-frame jitter effects.

Joint Weighting Mechanism. We define the distance between the Reference template (Standard Motion) $R$ and the Input sequence (User Motion) $I$ using a weighted Euclidean metric. While $\alpha$ and $\beta$ control the attention to dynamic derivatives \textit{within} a feature, the joint weight $w_k$ governs the spatial attention \textit{across} different body parts. Specifically, $w_k$ represents the importance of the $k$-th joint for the target action (e.g., high weights for knees in jumping, zero for wrists). The distance between the $i$-th frame of the reference $R_i$ and the $j$-th frame of the input $I_j$ is calculated as shown in Eq.~(\ref{eq:dist}):

\begin{equation}
	D(R_i, I_j) = \sqrt{\sum_{k=1}^{K} w_k \cdot (R_{i,k} - I_{j,k})^2}
	\label{eq:dist}
\end{equation}

Two-Stage Matching Strategy. To handle redundant actions in input videos, we propose a two-stage matching strategy that first performs subsequence search and then conducts keyframe alignment.By restricting the search space to a band with radius $r$ around the diagonal, FastDTW~\cite{21} reduces the time complexity from $O(N^2)$ to $O(N)$, enabling real-time processing of high-dimensional feature sequences.

Stage 1: Subsequence Matching. The goal is to locate the action interval $[t_{start}, t_{end}]$ in input $I$ that best matches reference $R$. A sliding-window search calculates the normalized FastDTW distance between the reference template and each windowed proposal. This effectively isolates the core action by finding the subsequence with the minimum alignment cost, discarding irrelevant pre- and post-action frames.

Stage 2: Keyframe Alignment via Warping Path. With the optimal interval fixed, we perform a fine-grained alignment to map reference keyframes $\mathcal{K}_{ref} = \{k_1, \dots, k_n\}$ to the input sequence. FastDTW seeks the optimal warping path $W = \{(i_1, j_1), \dots, (i_K, j_K)\}$ that minimizes the cumulative distance, subject to the search radius constraint as shown in Eq.~(\ref{eq:fastdtw}):

\begin{equation}
	W^* = \operatorname*{argmin}_{W} \sum_{(i,j) \in W} D(R_i, I_j) \quad \text{s.t.} \quad |i - j| \le r
	\label{eq:fastdtw}
\end{equation}
where $D(R_i, I_j)$ is the weighted Euclidean distance defined in Eq.~\ref{eq:dist}, and $r$ is the search radius. This constraint ensures the alignment path does not deviate excessively from the diagonal, preventing pathological warping while enabling precise transfer of biologically defined key moments to the user motion sequences.

\subsubsection{KISMAM.}In sports motion analysis, raw metrics like joint angles and spatiotemporal parameters are often unintuitive, and a single anomaly may arise from multiple technical issues. To bridge the gap between numerical features and interpretable diagnosis, we propose the KISMAM. By contrasting user motion features with established target models via thresholding, KISMAM aggregates probabilities to map multi-dimensional deviations into a set of interpretable assessment metrics, identifying core technical deficiencies and corresponding intervention strategies.

A comparative baseline was built using 100 adolescent sprinters (16-18 years, 100m: 10.31-14.00s), covering phases such as start, acceleration, maximum speed, deceleration, and standing long jump. For each key action $k$, statistical distributions of kinematic metrics (e.g., joint angle $\theta$, flight time $t_{flight}$) define normal ranges $[T_{\min}^k, T_{\max}^k]$.

Using MotionDTW-extracted keyframes, KISMAM computes the user's kinematic value $V_{user}^k$ at each key moment and contrasts it with the target model thresholds to determine deviation $D^k$ as shown in Eq.~(\ref{eq:deviation}):

\begin{equation}
	D^k = 
	\begin{cases} 
		T_{\min}^k - V_{\text{user}}^k, & \text{if } V_{\text{user}}^k < T_{\min}^k \\
		V_{\text{user}}^k - T_{\max}^k, & \text{if } V_{\text{user}}^k > T_{\max}^k \\
		0, & \text{otherwise}
	\end{cases}
	\label{eq:deviation}
\end{equation}

A kinematic metric anomaly ($D^k > 0$) may indicate one or multiple technical issues. We define a set $\mathcal{P} = {P_1, \dots, P_{n}}$ of $n$ common problems, from "Inappropriate start step length" to "center of mass too high at start." A mapping matrix $W \in \mathbb{R}^{K \times M}$ specifies the contribution of each metrics anomaly to each problem. For example, "flight time larger than reference" may correspond to $P_1$ or $P_2$, while "support leg tibia angle larger than reference" may indicate $P_9$ or $P_{11}$.

Since one anomaly can arise from multiple causes and one problem can affect multiple metrics, KISMAM uses a probability aggregation mechanism to estimate the occurrence likelihood of each problem and identify the core technical issues. For the $m$-th problem, its comprehensive score $S_m$ is calculated as in Eq.~(\ref{eq:score}):

\begin{equation}S_m = \sum_{k=1}^{K} D^k \cdot W_{km}\label{eq:score}\end{equation}

By normalizing the scores $S$ of all problems, KISMAM derives the probability distribution for each issue, ranking the set $\mathcal{P}$ and selecting the top $N$ (here $N=6$) as core diagnostic results. These highlight not only "what is wrong" but also the biomechanical rationale behind each problem. Unlike prior LLM-FMS approaches~\cite{33}, which feed raw angles or expert rules directly into LLMs-often causing hallucinations-this hybrid design assigns precise data processing and diagnosis, while the generative model handles semantic summarization. KISMAM thus bridges objective biomechanical characteristics and deterministic semantic diagnosis, enhancing both interpretability and system robustness.

\subsection{SportsRAG: RAG-based training guidance model}Although KISMAM effectively maps kinematic deviations into structured assessment metrics (e.g., \textit{insufficient start extension}), these numerical indicators lack contextual explanation and actionable guidance for end-users. Direct use of general Large Language Models (LLMs) often leads to domain knowledge gaps and severe hallucinations. To address this, we propose SportsRAG, a RAG-based training guidance model built upon Qwen3-8B. Unlike traditional fine-tuning approaches, SportsRAG leverages a massive external knowledge base to ground generation in verified expert data, ensuring scientific rigor and minimizing hallucinations.

To serve as the retrieval source for RAG, we construct a 6B-token knowledge base, categorized into three levels of granularity to ensure coverage and depth:

\begin{itemize}
	\item \textbf{Theoretical Foundation (Academic Corpus):} We curate over 200 authoritative textbooks (e.g., \textit{Sports Biomechanics}, \textit{Exercise Physiology}) and core sports science journals from the past decade, providing the theoretical bounds for generation.
	
	\item \textbf{Practical Experience (Expert QA):} To align with real-world needs, we aggregate common inquiries from expert daily training and rehabilitation routines, alongside queries from web users. Through rigorous manual annotation, we constructed a dataset of 50,000 high-quality QA pairs, serving as precise retrieval anchors for specific scenarios.
	
	\item \textbf{Reference Standards (Historical Reports):} We incorporate 1,000 professional analysis reports historically generated by student athletes using the AI system. These records exemplify the complete mapping from quantitative assessment data to specific training schemes and guidance, providing proven templates for reliable report generation.
\end{itemize}

A core innovation of SportsRAG is bridging the modality gap between quantitative metrics and textual knowledge. We implement a semantic mapping mechanism where a set of interpretable assessment metrics $X$ from KISMAM is used as a query key. The retrieval process operates in three sequential phases: first, via \textbf{Query Transformation}, the semantic labels corresponding to the detected deviations in $X$ (e.g., "knee valgus") are converted into dense metrics queries; next, \textbf{Context Retrieval} utilizes these queries to search the metrics database for the top-$K$ most relevant text chunks, encompassing expert explanations, historical cases, and intervention drills; finally, through \textbf{Knowledge Fusion}, this retrieved domain knowledge is concatenated with the KISMAM diagnosis to construct an augmented prompt for the LLM.

\section{Experiments}
\subsection{Performance on Motion Assessment}
To assess the practical value of precise keyframe extraction, we integrate our method into the downstream \textbf{KISMAM} assessment pipeline. We utilize the acceleration phase test set to extract keyframes via FastDTW (Baseline) and MotionDTW (Ours), which are then input into KISMAM to generate the Top-6 core biomechanical diagnosis. We evaluate performance using the Intersection over Union (IoU) metric. Let $\mathcal{D}$ denote the set of biomechanical diagnostic rules. Given the model's Top-6 prediction set $\mathcal{S}_{pred} \subset \mathcal{D}$ and the expert ground truth $\mathcal{S}_{gt} \subset \mathcal{D}$, the IoU is defined as shown in Eq.~(\ref{eq:iou}):

\begin{equation}
	\text{IoU} = \frac{|\mathcal{S}_{pred} \cap \mathcal{S}_{gt}|}{|\mathcal{S}_{pred} \cup \mathcal{S}_{gt}|}
	\label{eq:iou}
\end{equation}where $|\cdot|$ represents cardinality. This metric quantifies the ratio of correctly diagnosed rules (intersection) to the total unique rules identified (union).

\begin{figure}[!ht]
	\centering
	\includegraphics[width=0.6\textwidth]{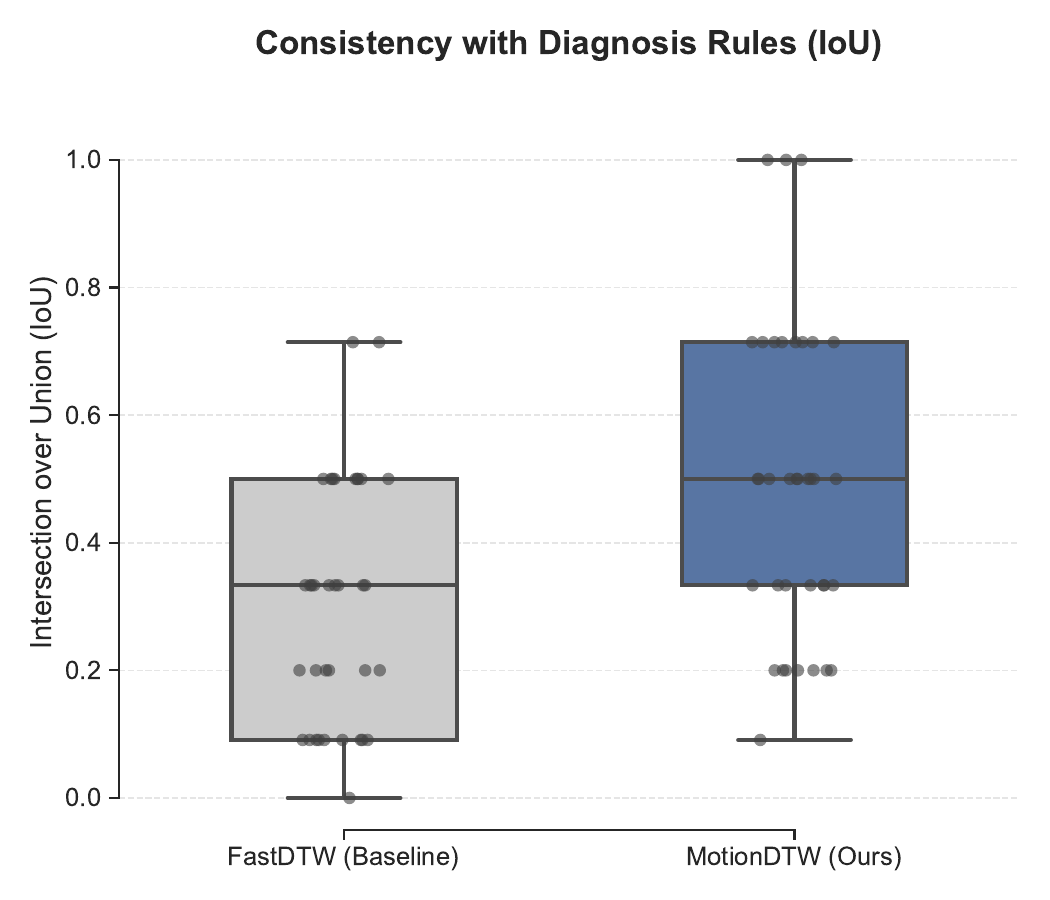} 
	\caption{\textbf{Downstream Diagnosis Consistency.}The box plot compares the Intersection over Union (IoU) of diagnosis rules generated by FastDTW and MotionDTW against ground truth.Our method (blue) shows significantly higher median IoU and upper-bound performance.}
	\label{fig:consistency}
\end{figure}

The results, visualized in Figure~\ref{fig:consistency}, demonstrate that MotionDTW significantly outperforms the baseline in preserving diagnostic semantic information. While the alignment error of \textbf{0.91 frames} implies that not all samples achieve perfect synchronization-resulting in IoU scores below 1.0 for some cases due to sensitive kinematic thresholds-our method exhibits a much stronger distribution. Notably, a substantial portion of our results reaches an IoU of \textbf{1.0}, indicating that MotionDTW can replicate expert-level diagnosis in scenarios where the baseline fails to capture critical kinematic features.

\subsection{Performance on Training Guidance}
To rigorously assess utility, we conduct a \textbf{double-blind evaluation} where three senior specialists score \textbf{60 randomly sampled reports} on a 1-5 Likert scale across four dimensions: \textbf{Accuracy} (interpretation correctness), \textbf{Comprehensiveness} (inference depth), \textbf{Professionalism} (expert-level specificity), and \textbf{Feasibility} (actionability of quantitative details).

To position our work within the current landscape of foundation models, we conduct a comparative study against four state-of-the-art generalist LLMs: \textbf{GPT-5}, \textbf{Claude 4.5 Opus}, \textbf{Gemini 3 Pro}, and \textbf{GLM 4.6V}. The quantitative results are listed in Table~\ref{tab:model_comparison}.

\begin{table}[!ht]
	\centering
	\caption{\textbf{Comparative Evaluation against Generalist LLMs.}}
	\label{tab:model_comparison}
	\begin{tabular}{|l|c|c|c|c|c|}
		\hline
		\textbf{Metric} & \textbf{SportsGPT} & \textbf{GPT-5} & \textbf{Claude 4.5} & \textbf{Gemini 3} & \textbf{GLM 4.6V} \\
		\hline
		Accuracy & \textbf{3.80} & 3.15 & 3.00 & 2.38 & 2.68 \\
		Comprehensiveness & \textbf{3.75} & 3.43 & 3.65 & 2.45 & 2.89 \\
		Professionalism & \textbf{3.73} & 3.07 & 3.13 & 2.70 & 2.35 \\
		Feasibility & \textbf{3.77} & 3.53 & 3.63 & 2.97 & 2.88 \\
		\hline
	\end{tabular}
\end{table}

As shown in Table~\ref{tab:model_comparison}, \textbf{SportsGPT} outperforms all baselines, peaking in \textbf{Accuracy} and \textbf{Feasibility}. While Claude 4.5 Opus shows competitive performance on professionalism (3.13), it still trails our specialized framework. Conversely, generalist models face challenges in maintaining high Accuracy. This performance drop suggests that current multi-modal alignment mechanisms lack the granularity required to precisely comprehend complex visual actions.

Expert feedback highlights three critical deficiencies in generalist models: (1) \textbf{Lack of Fine-Grained Perception:} They often treat motion data as abstract numbers, lacking the capability to detect subtle anomalies; (2) \textbf{Generic Templates:} Generalists tend to offer safe, "one-size-fits-all" advice (e.g., "maintain core stability") rather than targeted corrections; and (3) \textbf{Stochastic Instability:} They yield inconsistent diagnoses for identical inputs. In contrast, SportsGPT ensures stable, evidence-based assessments via its RAG-anchored architecture.

\subsection{Ablation Study}
\subsubsection{MotionDTW Ablation Study.}
To validate the robustness of our alignment algorithm, we calculate the \textbf{Average Keyframe Error} across all five test scenarios (four phases of the 100m sprint and the standing long jump). We systematically analyze the contribution of each module and the impact of the search strategy. The quantitative results are summarized in Table~\ref{tab:motiondtw_ablation}.

\textbf{Effectiveness of Functional Modules.}
As shown in the first row of Table~\ref{tab:motiondtw_ablation}, our \textbf{Full Model} achieves the lowest average error of \textbf{1.54 frames}, demonstrating superior alignment precision. Removing any individual component leads to performance degradation:

Removing dynamic features (angular velocity and acceleration) causes the error to nearly double to 2.74 frames, confirming that static joint angles alone are insufficient for high-speed sports and that higher-order derivatives are essential for capturing rapid temporal changes. Similarly, eliminating joint weights increases the error to 2.61 frames, indicating that prioritizing core kinematic chains-such as the lower limbs in sprinting-effectively suppresses noise from irrelevant body parts. Furthermore, relying solely on single-joint analysis results in a moderate error increase to 1.79 frames, suggesting that holistic body coordination is beneficial for robust representation.

\textbf{Necessity of the Two-Stage Strategy.}
A critical observation from Table~\ref{tab:motiondtw_ablation} is the catastrophic failure of the \textbf{One-Stage} strategy. Without the global coarse search to locate the action interval, the fine-grained alignment fails to converge within the long video sequence, resulting in an unacceptable average error of \textbf{66.00 to 76.81 frames}. This massive disparity (1.54 vs. 66.00) empirically proves that the Coarse-to-Fine Two-Stage mechanism is not merely an optimization, but a fundamental prerequisite for applying DTW to unconstrained sports videos.

\begin{table}[!ht]
	\centering
	\caption{\textbf{Impact of Alignment Strategies and Modules.} We report the average keyframe error (in frames) averaged across five motion categories (100m sprint phases and standing long jump). The Two-Stage strategy significantly outperforms the One-Stage approach, and the Full Model yields the best precision.}
	\label{tab:motiondtw_ablation}
	\resizebox{\textwidth}{!}{ 
		\begin{tabular}{|l|c|c|c|c|}
			\hline
			\textbf{Strategy} & \textbf{Full Model} & \textbf{w/o Joint Weights} & \textbf{w/o Dyn. Feat.} & \textbf{w/o Multi-Joint} \\
			\hline
			\textbf{Two-Stage (Ours)} & \textbf{1.54} & 2.61 & 2.74 & 1.79 \\
			One-Stage & 66.00 & 73.33 & 76.81 & 69.15 \\
			\hline
		\end{tabular}
	}
\end{table}

\begin{figure*}[t]
	\centering
	\includegraphics[width=\textwidth]{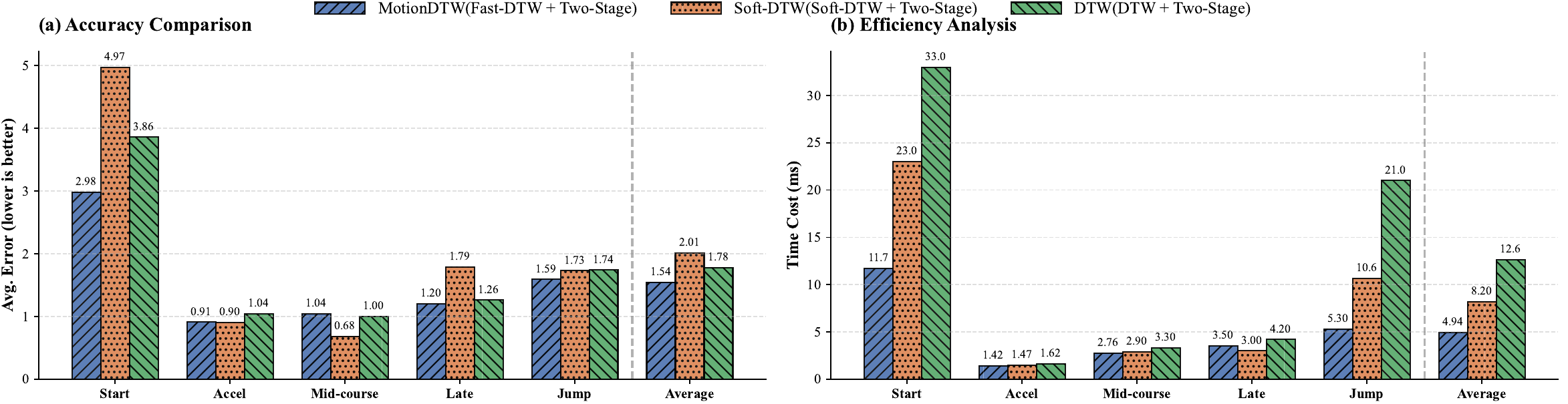}
	\caption{\textbf{Comprehensive Performance Analysis.} (a) Accuracy comparison showing MotionDTW achieving the lowest average error across all phases. (b) Efficiency analysis confirming real-time performance.}
	\label{fig:comparison}
\end{figure*}

We first validate the rationale behind building our framework upon FastDTW by benchmarking our \textbf{MotionDTW} against standard DTW\cite{19} and Soft-DTW\cite{38} across accuracy, efficiency, and stability metrics. As illustrated in Figure~\ref{fig:comparison} (a) and (b), while standard DTW offers theoretical exactness, its excessive computational overhead renders it prohibitive for real-time sports scenarios. Conversely, although Soft-DTW achieves competitive accuracy in specific phases (e.g., Mid-course), it exhibits critical instability. 

Notably, case-level analysis reveals that Soft-DTW is prone to "alignment collapse" during high-speed motion segments, resulting in frequent matching failures. In contrast, MotionDTW maintains a robust equilibrium between alignment precision and speed. It achieves the lowest average error of \textbf{0.91 frames} in the Acceleration phase while sustaining an ultra-low inference latency of \textbf{1.42 ms}, demonstrating superior stability for practical deployment.

\subsubsection{KISMAM Ablation Study.}
To validate the role of semantic translation, we compare our full framework against a variant where the KISMAM module was replaced by raw numeric thresholds (\textit{w/o KISMAM}). The results reveal a significant performance degradation in the absence of semantic mapping: \textbf{Accuracy drops markedly from 3.9 to 2.85}, and \textbf{Comprehensiveness declines from 3.85 to 2.4}. This indicates that without the structured assessment metrics provided by KISMAM, the LLM struggles to derive meaningful insights from raw biomechanical data alone. Consequently, it fails to identify complex, multi-joint coordination deficits, confirming that KISMAM acts as an essential "semantic bridge" between quantitative metrics and qualitative expert reasoning.
\subsubsection{SportsRAG Ablation Study.}
We further evaluated the necessity of external knowledge by ablating the Retrieval-Augmented Generation (RAG) module. The results highlight a critical functional separation: while \textbf{Accuracy} remains relatively robust at 3.65 (due to the retention of KISMAM for diagnosis), \textbf{Feasibility} suffers a catastrophic drop from \textbf{3.9 to 1.65}. Qualitative analysis reveals that without access to the expert knowledge base, the model's output regresses to theoretically correct but operationally vague suggestions (e.g., generic advice to "strengthen leg muscles" versus specific instructions for "4 sets of 8 reps at 85\% 1RM"). This empirical evidence confirms that RAG is a non-negotiable component for transforming diagnostic insights into executable, professional-grade training prescriptions.

\section{Conclusion and Future Work}

In this paper, we present an LLM-driven framework for interpretable sports motion assessment and training guidance that bridges the gap between quantitative biomechanical characteristics and actionable expert advice. By synergizing MotionDTW for precise spatiotemporal alignment, KISMAM for semantic diagnostic mapping, and a RAG-Based training guidance model, our system effectively mitigates the interpretability gap and hallucination issues prevalent in existing solutions. Empirical results confirm that our approach achieves state-of-the-art alignment precision and generates professional-grade training prescriptions comparable to those of human experts, offering a robust paradigm for intelligent coaching.

Looking ahead, we aim to enhance algorithmic robustness and ecosystem accessibility. We plan to upgrade MotionDTW by incorporating adaptive hyperparameter optimization and a multi-reference voting mechanism to handle extreme stylistic variations and mitigate the bias inherent in relying on a single reference sequence. Concurrently, we will expand our benchmark to cover diverse sporting disciplines (e.g., gymnastics, ball games) and release an open-source model zoo featuring SportsGPT variants of different scales, democratizing access to professional sports analysis. Finally, we plan to conduct longitudinal intervention studies to empirically validate the system's practical efficacy in real-world athletic training scenarios.

\bibliographystyle{splncs04}
\bibliography{refs}

\end{document}